%% file: main.tex
\pdfoutput=1
\documentclass[11pt]{article}
\PassOptionsToPackage{table}{xcolor}

\usepackage[preprint]{acl}
\usepackage{times}
\usepackage{latexsym}

\usepackage[T1]{fontenc}

\usepackage[utf8]{inputenc}

\usepackage{microtype}

\usepackage{inconsolata}

\usepackage{graphicx}

\usepackage{booktabs}
\usepackage{xcolor}
\usepackage{multirow}
\usepackage{makecell}
\usepackage{amsmath}
\usepackage{array}

\usepackage{amssymb}

\usepackage{amsthm}

\hypersetup{
  pdftitle={Rethinking Reverse KL as Adaptive Entropy Distillation},
  pdfauthor={Shizhen Li et al.},
  pdfsubject={Adaptive Entropy Distillation for knowledge distillation},
  pdfkeywords={knowledge distillation, reverse KL, large language models}
}

\title{Rethinking Reverse KL as Adaptive Entropy Distillation}

\author{
Shizhen Li\textsuperscript{1},
Zhiyu Shen\textsuperscript{1},
Yuyin Lu\textsuperscript{1},
Yunhe Pang\textsuperscript{1},
Jielin Song\textsuperscript{1},
Yanghui Rao\textsuperscript{1}\thanks{Corresponding author},
Fu Lee Wang\textsuperscript{2}
\\
\textsuperscript{1}School of Computer Science and Engineering, 
Sun Yat-sen University, Guangzhou, China
\\
\textsuperscript{2}School of Science and Technology, 
Hong Kong Metropolitan University, Hong Kong SAR, China
\\
\texttt{\{lishzh57, shenzhy23, luyy37, pangyh8, songjlin6\}@mail2.sysu.edu.cn}
\\
\texttt{raoyangh@mail.sysu.edu.cn \quad pwang@hkmu.edu.hk}
}

\begin{document}
\maketitle
\begin{abstract}

Knowledge distillation (KD) is widely used to transfer the capabilities of large language models (LLMs) to smaller students, but existing objectives often struggle to balance faithful imitation and robust generation. 
In particular, existing methods mainly combine FKL and RKL, overlooking that RKL itself provides a mechanism for adjusting the student's imitation strength.
Motivated by this, we revisit on-policy Reverse Kullback-Leibler (RKL) distillation and decompose its objective into a teacher-fitting term and a student-entropy term, without introducing an explicit FKL branch. We show theoretically that the token-level optimal student distribution corresponds to a tempered variant of the teacher distribution, where the adaptive weight controls the trade-off between mode-seeking and uncertainty preservation. Guided by this insight, we propose \textbf{Adaptive Entropy Distillation (AED)}, which uses the teacher's entropy to dynamically calibrate token-level imitation strength. Experiments on instruction-following and mathematical reasoning benchmarks demonstrate that AED achieves superior overall performance and generally improves teacher--student distributional and entropy alignment.\footnote{Code: \href{https://github.com/ShizhenL1/AED}{\texttt{https://github.com/ShizhenL1/AED}}}

\end{abstract}

\input{chap1_intro}
\input{chap2_relate}

\input{chap3_pre}

\input{chap4_method}

\input{chap5_ex}

\input{chap6_conclu}
\input{chap7_limit}

\bibliography{main}

\input{chap8_app}

\end{document}

%% file: chap1_intro.tex
\section{Introduction}

\begin{figure}[t]
  \includegraphics[width=\columnwidth]{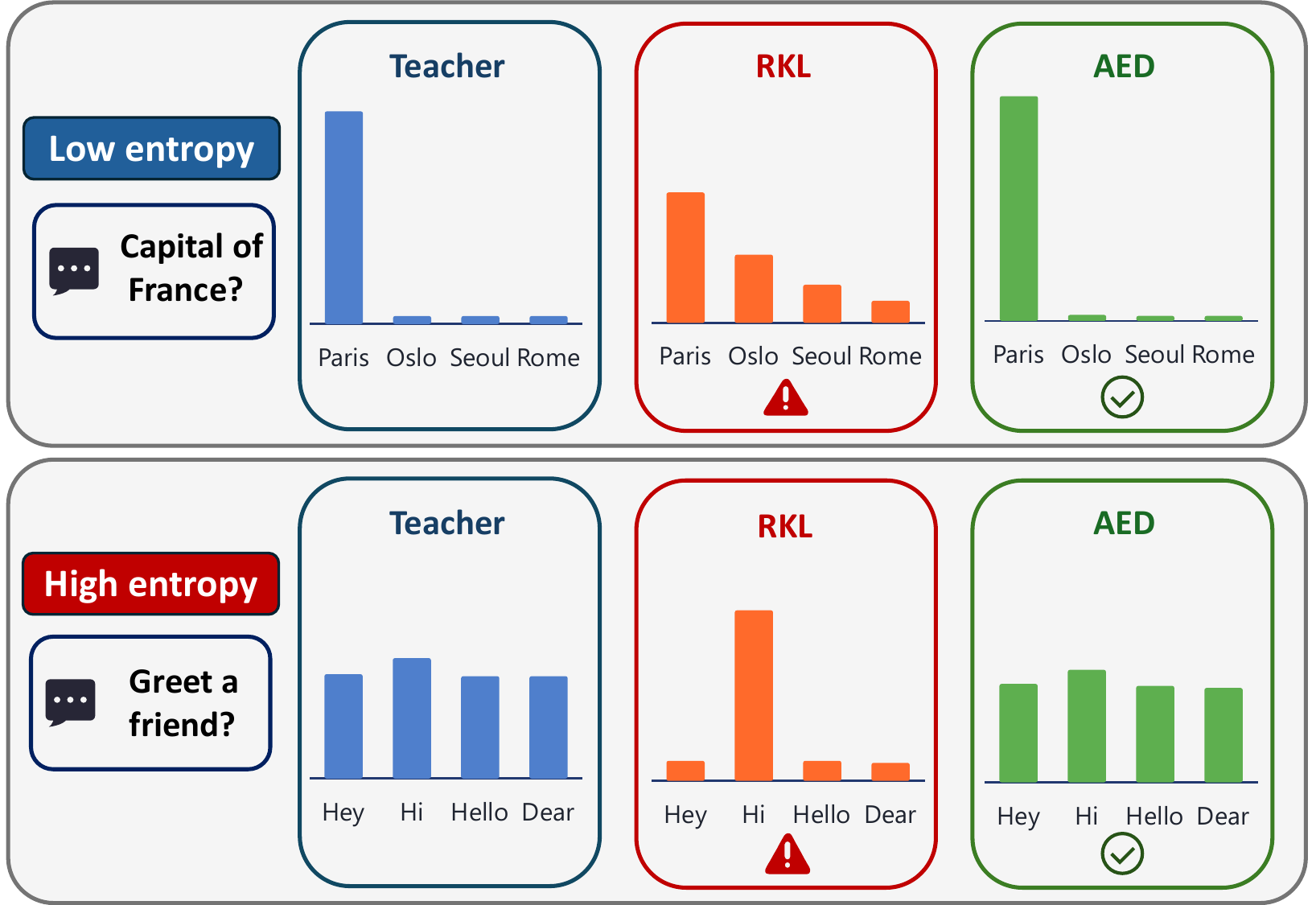}
  \caption{Comparison between fixed-strength distillation and AED: fixed strength applies uniform imitation, while AED adapts to teacher uncertainty.}
  \label{fig:first}
\end{figure}

LLMs offer strong capabilities but incur high computational and deployment costs. KD \citep{kd} addresses this issue by compressing large teacher models into lightweight students. Conventional sequence-level KD \citep{DBLP:conf/emnlp/KimR16} optimizes FKL divergence on teacher-generated trajectories, but such off-policy training suffers from exposure bias \citep{DBLP:conf/nips/BengioVJS15, DBLP:journals/corr/RanzatoCAZ15}. 
Recent work turns to on-policy RKL distillation with student-sampled trajectories \citep{minillm}, reducing the training-inference mismatch.

Nevertheless, fully exploiting the benefits of on-policy distillation remains challenging due to the intrinsic trade-off between the probability coverage of FKL and the probability concentration of RKL \citep{minillm}. Existing frameworks often address this through static interpolations \citep{gkd} or variants like Skewed RKL \citep{distillm}. Yet, these global, fixed-weighting strategies uniformly scale objectives across entire sequences, failing to account for varying token-level teacher confidence. To overcome this, recent methods such as AKL \citep{akl} and ToDi \citep{todi} have introduced dynamic balancing strategies to avoid uniform scaling via token-level mechanisms.

While existing frameworks mainly focus on adaptively combining FKL and RKL, the internal structure of RKL itself already provides a mechanism for controlling the strength of the student's imitation of the teacher.

We therefore revisit on-policy distillation from the perspective of distribution geometry. We show that the RKL objective itself contains complementary forces for concentration and coverage, enabling an RKL-only formulation by reweighting its two internal components. Under this reweighted formulation, the analytical optimum corresponds to a tempered teacher distribution, providing a theoretical characterization of how the student's imitation strength can be controlled.

Building on these theoretical findings, we develop AED, a theory-driven, uncertainty-aware framework for on-policy distillation, as illustrated in Figure~\ref{fig:first}. AED is motivated by this theoretical optimum and adaptively adjusts the imitation strength at the token level.
To balance the two terms within the RKL, AED introduces an adaptive weighting mechanism. We instantiate this mechanism by mapping the teacher's entropy to the weight, encouraging the student internalizes the teacher's distribution.
We evaluate AED on instruction distillation and mathematical reasoning benchmarks and further analyze its token-level distributional behavior.

Grounded in a geometric analysis of RKL, AED reweights its intrinsic components to yield an RKL-only objective and uses teacher entropy to calibrate token-level imitation strength. Experiments demonstrate that this design improves task performance and teacher--student distributional alignment, highlighting the practical value of the theoretical characterization.

%% file: chap2_relate.tex
\section{Related Work}
\label{sec:related_work}

\subsection{Evolution of LLM KD}
Traditional KD is a major model compression technique, typically following an off-policy paradigm where the student learns from static teacher-generated targets, such as soft logits, intermediate representations, or rationales \cite{kd, DBLP:journals/corr/abs-1910-01108, DBLP:conf/emnlp/JiaoYSJCL0L20}. For LLMs, it has been used to distill reasoning via chain-of-thought rationales \cite{DBLP:conf/acl/HsiehLYNFRKLP23, DBLP:conf/icml/FuPOSK23} and instruction-following behaviors from synthetic datasets \cite{DBLP:conf/emnlp/JiangCCW23}. However, off-policy KD suffers from exposure bias and cumulative distribution shift \cite{DBLP:conf/nips/BengioVJS15, DBLP:journals/corr/RanzatoCAZ15}, since the student is not trained on states induced by its own autoregressive errors.

To address this, recent on-policy distillation methods let the student sample from its own distribution while receiving teacher supervision \cite{minillm, gkd}. MiniLLM \cite{minillm} uses RKL divergence for token-level corrective signals, while GKD \cite{gkd} generalizes on-policy distillation with alternative divergence objectives. However, these methods often treat RKL as a fixed objective; we further decompose and analyze RKL to better understand its distillation behavior.

\subsection{The FKL-RKL Trade-off in Generative Modeling}
The choice of divergence measure strongly shapes the student's behavior. Prior work has explored alternative objectives for on-policy distillation, including Skewed RKL in DistilLLM \cite{distillm}, JS divergence in GKD \cite{gkd}, SKL and general $f$-divergences \cite{f-diver}. These objectives aim to balance the mode-seeking behavior of RKL with broader distributional coverage.

Beyond static divergence selection, recent work further explores adaptive balancing strategies. Related adaptive weighting methods have been studied in computer vision \citep{DBLP:conf/iclr/ZhengY24, DBLP:journals/corr/abs-2212-12965}, while autoregressive sequence generation emphasizes token-level dynamics for balancing different divergence behaviors. 
AKL \cite{akl} adapts FKL and RKL based on their different fitting behaviors over the head and tail regions of the teacher distribution, whereas ToDi \cite{todi} coordinates them according to how their gradients increase or decrease token probabilities during on-policy sampling. More recently, EOPD \cite{eopd} builds on RKL and introduces an additional FKL term for high-entropy teacher tokens.
In contrast, AED does not interpolate FKL and RKL objectives or activate an additional FKL branch. Instead, it exploits the distinct fitting roles of RKL's two internal terms, yielding an adaptive target distribution and a simple RKL-only on-policy distillation objective.

%% file: chap3_pre.tex
\section{Preliminaries and Analysis}
This section first introduces the problem setup, notation, and KL-based distillation objectives, and then examines the training behaviors induced by these objectives, which in turn motivate the adaptive method developed later.

\subsection{Problem Setup and Notation}
Let $P$ denote the teacher model and $Q_{\theta}$ denote the student model parameterized by $\theta$.
Given an input prompt $x$ and an output sequence
$y=(y_1,\dots,y_T)$, the teacher and student define autoregressive sequence distributions as
\begin{equation}
\begin{aligned}
p(y \mid x)
&=
\prod_{t=1}^{T}
p(y_t \mid x, y_{<t}),
\\
q_{\theta}(y \mid x)
&=
\prod_{t=1}^{T}
q_{\theta}(y_t \mid x, y_{<t}),
\end{aligned}
\end{equation}
where $y_{<t}=(y_1,\dots,y_{t-1})$ denotes the prefix preceding step $t$.

In on-policy distillation, response are sampled from the student model $y \sim q_{\theta}(\cdot \mid x)$. For a given prefix $y_{<t}$, we denote the teacher and student conditional distributions over the entire vocabulary $\mathcal{V}$ at step $t$ as $p_t$ and $q_t$, respectively. Specifically, for any token $v \in \mathcal{V}$:
\begin{equation}
    p_t(v) = p(v \mid x, y_{<t}), \quad q_t(v) = q_\theta(v \mid x, y_{<t}).
\end{equation}
Under this notation, sequence-level objectives can be decomposed into token-level terms along the sampled trajectory.

\subsection{Geometric Analysis of KL Distillation}

The different behaviors of forward and reverse KL can be understood through the training objectives they induce.
To analyze the optimization dynamics in the on-policy setting, we reformulate the sequence-level objectives into token-level expectations over student-generated trajectories. 
For RKL, minimizing
$D_{\mathrm{KL}}(q_{\theta} \parallel p)$
is equivalent to minimizing the following on-policy objective over responses sampled from the student model:
\begin{equation}
    \mathcal{L}_{\mathrm{RKL}}
    =
    \mathbb{E}_{y \sim q_\theta(\cdot|x)}
    \left[
    \sum_{t=1}^T \sum_{v \in \mathcal{V}}
    q_t(v)
    \log \frac{q_t(v)}{p_t(v)}
    \right].
\end{equation}
This objective reveals two competing forces in training. 
The log-ratio naturally separates into $-\sum_{v \in \mathcal{V}} q_t(v)\log p_t(v)$ and $\sum_{v \in \mathcal{V}} q_t(v)\log q_t(v)$. 
The first term, $-\sum_{v \in \mathcal{V}} q_t(v) \log p_t(v)$, 
acts as a teacher-derived cross-entropy loss. When minimized, it heavily penalizes the student for assigning non-negligible probability mass to tokens that the teacher deems unlikely, thereby driving the student distribution toward the teacher's high-probability modes.
The second term, $\sum_{v \in \mathcal{V}} q_t(v) \log q_t(v)$, corresponds to the negative entropy of the student distribution, which inherently penalizes over-concentration.

As a consequence, RKL-based training can be interpreted as balancing teacher-driven mode seeking against entropy-induced probability spreading on student-sampled trajectories.

Similarly, while standard FKL is off-policy (i.e., evaluated on teacher-generated sequences), recent on-policy distillation methods often employ a token-level FKL formulation to mitigate exposure bias \citep{gkd}. In this setting, prefixes are sampled from the student model, and the student is forced to match the teacher's vocabulary distribution at each step via token-level FKL. The corresponding on-policy FKL objective can be written as:
\begin{equation}
    \mathcal{L}_{\text{On-FKL}}
    =
    \mathbb{E}_{y \sim q_\theta(\cdot|x)}
    \left[
    \sum_{t=1}^T \sum_{v \in \mathcal{V}}
    p_t(v)
    \log \frac{p_t(v)}{q_t(v)}
    \right].
\end{equation}
By separating the log-ratio, the token-level objective can be decomposed into two terms: $-\sum_{v \in \mathcal{V}} p_t(v)\log q_t(v)$ and $\sum_{v \in \mathcal{V}} p_t(v)\log p_t(v)$.
In this decomposition, the first term,
$-\sum_{v \in \mathcal{V}} p_t(v) \log q_t(v)$
is the cross-entropy term.
Under the teacher-weighted objective, whenever the student assigns insufficient probability mass to teacher-supported tokens, this term induces large corrective gradients and therefore forces the student to cover the support of the teacher distribution as broadly as possible.

In contrast, the second term, $\sum_v p_t(v)\log p_t(v)$, equals $-H(p)$ and depends only on the teacher distribution, making it constant with respect to the student parameters.
It therefore contributes no gradient to student optimization.
As a result, FKL-based training is driven entirely by the support-covering pressure of the cross-entropy term, which may lead to mean-seeking behavior when the student family is less expressive than the teacher.

RKL, however, has a different internal structure. The decomposition above shows that it already contains two complementary terms: a teacher-fitting term and a student-entropy term. In standard RKL, these two terms have equal relative weights. This naturally raises the following question: \textbf{what happens if we explicitly re-weight them?}
Increasing the weight of the teacher-fitting term sharpens
the student distribution toward the teacher's
high-probability regions, whereas increasing the weight of
the student-entropy term produces a smoother distribution
with broader probability coverage. This suggests that the
trade-off between probability concentration and coverage
can be controlled directly through the intrinsic components
of RKL, motivating the adaptive re-weighting formulation
introduced in the next section.

%% file: chap4_method.tex
\begin{figure*}[ht]
  \includegraphics[width=\linewidth]{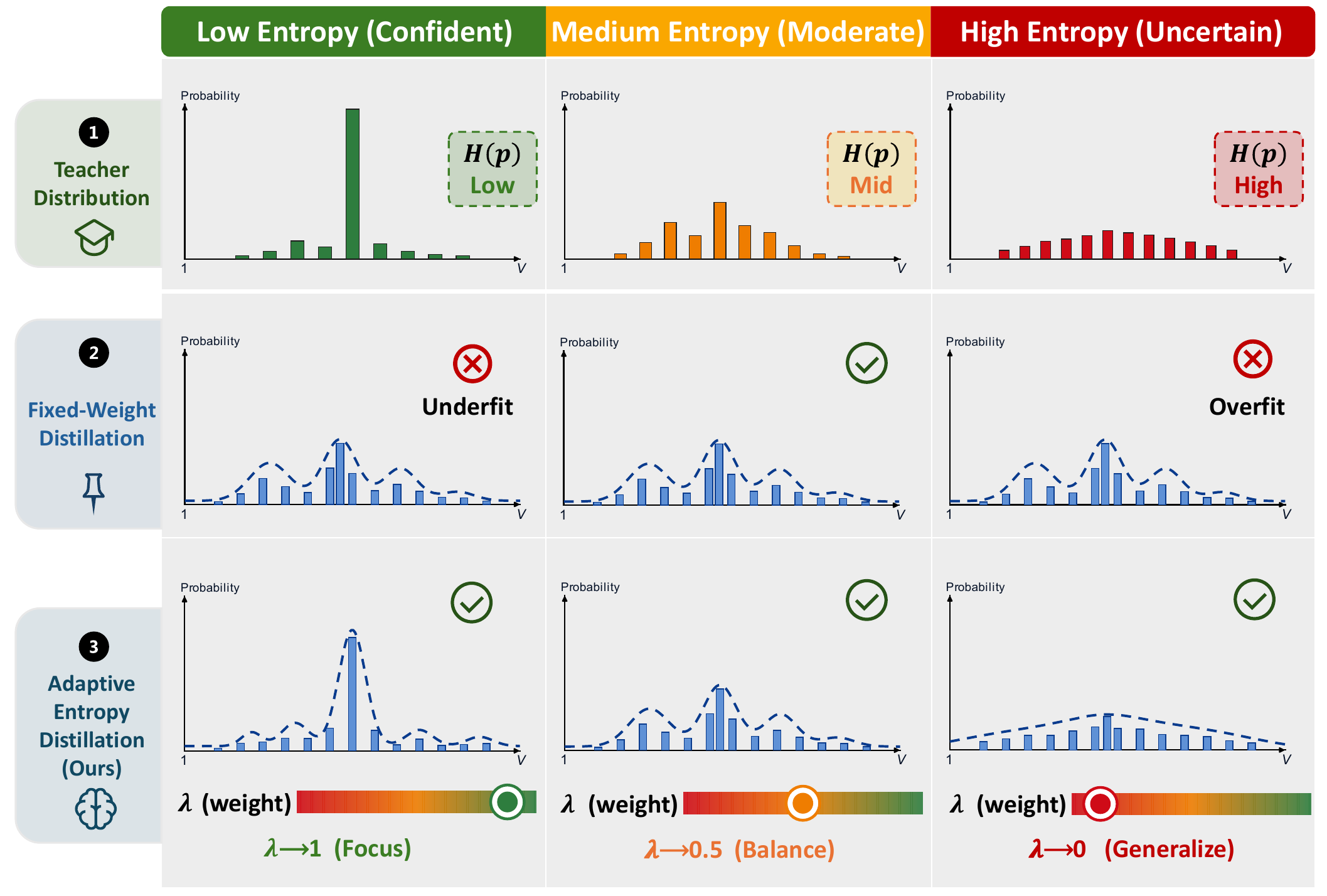}
  \caption{Conceptual comparison between fixed-weight distillation and AED.}
  \label{fig:frame}
\end{figure*}

\section{Method}
In this section, we present AED, a novel on-policy distillation framework. We begin by deriving the analytical optimum of our adaptive objective, establishing its formal equivalence to the RKL divergence to a tempered teacher distribution. Guided by this theoretical insight, we introduce an entropy-driven mechanism that dynamically modulates the imitation strength at the token level, enabling the student to accurately internalize the teacher's distributional structure.

\subsection{Analytic Optimum and Equivalence to Tempered RKL}

To provide a theoretical characterization of AED, we analyze
the token-level adaptive loss $\mathcal{L}_t$ at a fixed
student-generated prefix. In this local analysis, the teacher
distribution $p_t$ and the adaptive weight $\lambda_t$ are
treated as fixed. The objective is defined as
\begin{equation}
\begin{aligned}
    \mathcal{L}_t(q_t)
    ={}& \sum_{v \in \mathcal{V}} q_t(v)
    \Big[
    -\lambda_t \log p_t(v) \\
    &\quad + (1-\lambda_t)\log q_t(v)
    \Big].
\end{aligned}
\label{equ:token_loss}
\end{equation}

\paragraph{Optimal student distribution.}
For a given teacher distribution $p_t$ and
$0<\lambda_t<1$, minimizing
Equation~\ref{equ:token_loss} over the probability simplex
yields
\begin{equation}
    q_t^*(v)
    =
    \frac{[p_t(v)]^{\alpha_t}}
    {\sum_{u\in\mathcal{V}}[p_t(u)]^{\alpha_t}},
    \qquad
    \alpha_t
    =
    \frac{\lambda_t}{1-\lambda_t}.
    \label{eq:optimal_distribution}
\end{equation}
The boundary cases can be understood through their
corresponding limits. The derivation is provided in
Appendix~\ref{app:analytic_optimum1}.

Equation~\ref{eq:optimal_distribution} shows that the
optimizer is a tempered variant of the teacher distribution,
with $\lambda_t$ directly controlling the sharpness of the
imitation target. To further elucidate the relationship between our objective and classical divergence measures, we derive the following equivalence:

\paragraph{Equivalent tempered-RKL form.}
To further clarify the relation between the adaptive
objective and conventional divergence minimization, define
the normalized tempered teacher distribution as
\begin{equation}
    \tilde{p}_t^{(\alpha_t)}(v)
    =
    \frac{[p_t(v)]^{\alpha_t}}
    {Z_t(\alpha_t)},
    \qquad
    Z_t(\alpha_t)
    =
    \sum_{u\in\mathcal{V}}[p_t(u)]^{\alpha_t}.
    \label{eq:tempered_teacher}
\end{equation}
The adaptive loss can then be rewritten as
\begin{equation}
    \mathcal{L}_t(q_t)
    =
    (1-\lambda_t)
    \left[
    D_{\mathrm{KL}}
    \left(
    q_t
    \parallel
    \tilde{p}_t^{(\alpha_t)}
    \right)
    -
    \log Z_t(\alpha_t)
    \right].
    \label{eq:tempered_equivalence}
\end{equation}
The corresponding derivation is provided in
Appendix~\ref{app:analytic_optimum1}.

Since $1-\lambda_t$ is positive and $\log Z_t(\alpha_t)$ is independent of $q_t$, minimizing the adaptive loss with respect to $q_t$ at a fixed prefix is equivalent to minimizing the RKL divergence to the corresponding tempered teacher distribution.

As AED averages over prefixes sampled from $q_\theta$, it induces a prefix-dependent conditional token-level target throughout training.

Equation~\ref{eq:tempered_equivalence} demonstrates that AED implements a controllable mode-seeking paradigm within the RKL framework.In this context, $\alpha_t$ serves as the inverse temperature: a high $\alpha_t$ (driven by a large $\lambda_t$) forces the student toward the teacher's primary modes, whereas a low $\alpha_t$ encourages broader probability coverage.

\subsection{AED}
\label{sec:aed_method}
Building upon the theoretical foundation established in equation~\ref{eq:optimal_distribution}, equation~\ref{eq:tempered_equivalence}, and the detailed derivation in Appendix~\ref{app:analytic_optimum1}, we introduce AED, as illustrated in Figure~\ref{fig:frame}. The core principle of AED is to dynamically modulate the balance between the teacher-fitting objective and the student-entropy, thereby enabling the student to adapt its imitation strength according to the teacher's local distributional characteristics.

As demonstrated in Eq.~(\ref{eq:tempered_equivalence}), the adaptive weight $\lambda_t$ serves as the critical controller for the inverse temperature $\alpha_t$, which governs the sharpness of the imitation target. To operationalize this, we define the sequence-level on-policy distillation loss for a given input $x$ as:
\begin{equation}
    \mathcal{L}_{\text{AED}}(x) = \mathbb{E}_{y \sim q_{\theta}(\cdot|x)} \left[ \sum_{t=1}^{T} \mathcal{L}_t(q_t) \right],
\end{equation}
where $\mathcal{L}_t(q_t)$ denotes the token-level adaptive loss defined in Eq.~(\ref{equ:token_loss}).

We formalize this uncertainty-aware calibration by mapping the teacher's token-level entropy $H_t(p)$ to an adaptive weight $\lambda_t \in [0, 1]$ via a generalized mapping function $f(\cdot)$:
\begin{equation}
    \lambda_t = f\big(H_t(p)\big).
\end{equation}
Specifically, the mapping function $f(\cdot)$ must satisfy two fundamental constraints to ensure valid distillation dynamics: (i) the output must be bounded within the interval $[0, 1]$ to serve as a proper weighting factor, and (ii) it must be monotonically decreasing with respect to the teacher's entropy $H_t(p)$, ensuring that the adaptive weight systematically decays as the teacher's uncertainty increases.

To avoid introducing additional mapping hyperparameters, we instantiate $f(\cdot)$ using the normalized divergence between the teacher distribution and the uniform distribution $U(v)=1/|V|$. Specifically,
\begin{equation}
\begin{aligned}
    D_{\mathrm{KL}}(p_t \parallel U)
    &= \sum_{v \in V} p_t(v)
    \left(\log p_t(v) + \log |V|\right) \\
    &= \log |V| - H_t(p).
\end{aligned}
    \label{eq:entropy_kl_divergence}
\end{equation}
We therefore set
\begin{equation}
    \lambda_t
    = \frac{D_{\mathrm{KL}}(p_t \parallel U)}{\log |V|}
    = 1 - \frac{H_t(p)}{\log |V|}.
    \label{eq:entropy_kl_mapping}
\end{equation}

Through this entropy-driven coupling, AED adaptively adjusts the imitation strength: when the teacher has low entropy, $\lambda_t$ approaches 1 and the student receives stronger guidance from the teacher. For high-entropy tokens, a smaller $\lambda_t$ reduces the imitation strength and increases the contribution of the student-entropy term.

Under limited student capacity and the geometry of the optimization landscape, the student may not always be able to directly reach the teacher distribution. By adapting imitation strength to teacher entropy, AED provides a more suitable optimization trajectory that helps the student recover the teacher's most informative distributional structure.

Importantly, AED begins by decomposing RKL and adaptively balancing its two terms according to teacher entropy. The tempered teacher distribution in Equation~\ref{eq:optimal_distribution} provides an equivalent characterization of the resulting target, rather than serving as an initial design assumption or a separately introduced temperature-scaling objective. In practice, AED directly optimizes Equation~\ref{equ:token_loss}.

\definecolor{lightgray}{RGB}{225,225,225}
\definecolor{lightblue}{RGB}{210,230,245}

\begin{table*}[h]
\centering
\small
\setlength{\tabcolsep}{9pt}
\renewcommand{\arraystretch}{1.15}

\begin{tabular}{lccccc|c}
\toprule
Methods & Dolly & SelfInst & Vicuna & S-NI & UnNI & Avg \\
\midrule

\multicolumn{7}{c}{GPT2 1.5B $\rightarrow$ GPT2 120M} \\
\midrule

\rowcolor{lightgray}
Teacher & 27.18$_{\pm0.18}$ & 15.07$_{\pm0.22}$ & 16.14$_{\pm0.46}$ & 27.46$_{\pm0.16}$ & 31.53$_{\pm0.14}$ & 23.48 \\

SFT & 18.74$_{\pm0.20}$ & 9.43$_{\pm0.56}$ & 11.92$_{\pm0.45}$ & 16.88$_{\pm0.37}$ & 18.54$_{\pm0.06}$ & 15.10 \\
\midrule

RKL
& 24.94$_{\pm0.31}$
& 10.12$_{\pm0.43}$
& \underline{16.29$_{\pm0.29}$}
& 17.97$_{\pm0.16}$
& 21.15$_{\pm0.10}$
& 18.09 \\

AKL
& 24.92$_{\pm0.31}$
& 10.36$_{\pm0.26}$
& 15.44$_{\pm0.49}$
& 18.41$_{\pm0.14}$
& 21.07$_{\pm0.07}$
& 18.04 \\

ToDi
& \underline{24.95$_{\pm0.19}$}
& 10.38$_{\pm0.34}$
& 15.22$_{\pm0.64}$
& 17.12$_{\pm0.13}$
& 19.70$_{\pm0.17}$
& 17.47 \\

EOPD
& 23.66$_{\pm0.34}$
& \textbf{12.66$_{\pm0.67}$}
& 15.74$_{\pm0.50}$
& \underline{23.52$_{\pm0.34}$}
& \underline{24.21$_{\pm0.12}$}
& \underline{19.96} \\

\rowcolor{lightblue}
AED
& \textbf{25.44$_{\pm0.41}$}
& \underline{12.61$_{\pm0.25}$}
& \textbf{16.83$_{\pm0.20}$}
& \textbf{25.33$_{\pm0.13}$}
& \textbf{27.74$_{\pm0.06}$}
& \textbf{21.59} \\

\midrule
\multicolumn{7}{c}{LLaMA2 7B $\rightarrow$ TinyLLaMA 1.1B} \\
\midrule

\rowcolor{lightgray}
Teacher & 29.59$_{\pm0.25}$ & 21.18$_{\pm0.28}$ & 20.58$_{\pm0.30}$ & 32.71$_{\pm0.16}$ & 32.72$_{\pm0.08}$ & 27.36 \\

SFT & 22.59$_{\pm0.31}$ & 16.39$_{\pm0.58}$ & 16.13$_{\pm0.36}$ & 26.65$_{\pm0.14}$ & 28.06$_{\pm0.15}$ & 21.96 \\
\midrule

RKL
& 25.23$_{\pm0.28}$
& \underline{17.94$_{\pm0.59}$}
& 17.03$_{\pm0.37}$
& \underline{29.04$_{\pm0.17}$}
& \underline{30.54$_{\pm0.13}$}
& \underline{23.96} \\

AKL
& 25.20$_{\pm0.19}$
& 17.31$_{\pm0.31}$
& 16.52$_{\pm0.62}$
& 27.53$_{\pm0.21}$
& 29.56$_{\pm0.19}$
& 23.22 \\

ToDi
& \underline{25.31$_{\pm0.50}$}
& 17.32$_{\pm0.29}$
& \underline{17.36$_{\pm0.60}$}
& 26.98$_{\pm0.27}$
& 29.44$_{\pm0.08}$
& 23.28 \\

EOPD
& 22.53$_{\pm0.24}$
& 15.16$_{\pm0.76}$
& 15.62$_{\pm0.31}$
& 25.30$_{\pm0.40}$
& 27.29$_{\pm0.17}$
& 21.18 \\

\rowcolor{lightblue}
AED
& \textbf{27.28$_{\pm0.23}$}
& \textbf{18.93$_{\pm0.31}$}
& \textbf{18.52$_{\pm0.20}$}
& \textbf{30.98$_{\pm0.27}$}
& \textbf{32.79$_{\pm0.13}$}
& \textbf{25.70} \\

\midrule
\multicolumn{7}{c}{Qwen3-4B $\rightarrow$ Qwen3-0.6B} \\
\midrule

\rowcolor{lightgray}
Teacher
& 31.08$_{\pm0.53}$
& 25.82$_{\pm0.32}$
& 23.05$_{\pm0.76}$
& 42.51$_{\pm0.30}$
& 39.00$_{\pm0.09}$
& 32.29 \\

SFT
& 26.61$_{\pm0.20}$
& 18.74$_{\pm0.59}$
& 18.57$_{\pm0.51}$
& 35.77$_{\pm0.42}$
& 35.03$_{\pm0.21}$
& 26.94 \\
\midrule

RKL
& 28.33$_{\pm0.30}$
& 20.10$_{\pm0.52}$
& 19.76$_{\pm0.60}$
& 36.94$_{\pm0.27}$
& 36.90$_{\pm0.05}$
& 28.41 \\

AKL
& 28.68$_{\pm0.29}$
& \textbf{21.16$_{\pm0.44}$}
& 20.29$_{\pm0.24}$
& \underline{38.29$_{\pm0.28}$}
& 37.69$_{\pm0.12}$
& \underline{29.22} \\

ToDi
& \textbf{29.09$_{\pm0.20}$}
& \underline{20.74$_{\pm0.45}$}
& 19.75$_{\pm0.36}$
& 36.85$_{\pm0.08}$
& \underline{37.90$_{\pm0.07}$}
& 28.86 \\

EOPD
& 25.55$_{\pm0.19}$
& 17.05$_{\pm0.65}$
& \textbf{22.24$_{\pm0.19}$}
& 32.59$_{\pm0.29}$
& 34.13$_{\pm0.10}$
& 26.31 \\

\rowcolor{lightblue}
AED
& \underline{28.99$_{\pm0.18}$}
& 20.01$_{\pm0.33}$
& \underline{20.96$_{\pm0.24}$}
& \textbf{38.69$_{\pm0.19}$}
& \textbf{39.10$_{\pm0.11}$}
& \textbf{29.55} \\

\bottomrule
\end{tabular}

\caption{ROUGE-L results on five instruction-following benchmarks. Results are reported as mean $\pm$ standard deviation across random seeds. \textbf{Bold} and \underline{underlined} values indicate the best and second-best results within each model configuration, respectively, among the five distillation methods.}
\label{tab:main_results}
\end{table*}

%% file: chap5_ex.tex
\section{Experiments}

\subsection{Experimental Setup}

\paragraph{Training Data and Models.} 
Following \citet{minillm}, we train on \texttt{databricks/dolly-15k}. We consider three teacher-student configurations: (1) GPT2: GPT-2 1.5B distilling to GPT-2 120M \citep{Radford2019LanguageMA}; (2) LLaMA: LLaMA2-7B \citep{DBLP:journals/corr/abs-2307-09288} distilling to TinyLLaMA-1.1B \citep{DBLP:journals/corr/abs-2401-02385}; (3) Qwen: Qwen3-4B \citep{qwen3} distilling to Qwen3-0.6B \citep{qwen3}.
Further implementation details and hyperparameters are provided in Appendix~\ref{app:imple_details}.

\paragraph{Evaluation Benchmarks and Metrics.}
Following the evaluation protocol of \citet{minillm}, we evaluate AED on five instruction-following benchmarks: \textbf{Dolly}, \textbf{SelfInst} \citep{DBLP:conf/acl/WangKMLSKH23}, \textbf{Vicuna} \citep{DBLP:conf/nips/ZhengC00WZL0LXZ23}, \textbf{S-NI} \citep{wang2022super}, and \textbf{UnNI} \citep{DBLP:conf/acl/HonovichSLS23}. Evaluation details are provided in Appendix~\ref{app:imple_details}. We use \textbf{ROUGE-L} \citep{lin2004rouge} as the primary metric and further conduct LLM-as-a-Judge pairwise preference evaluation to assess model alignment.

\paragraph{Baselines.}
We compare AED with the standard RKL baseline and three recent adaptive distillation methods: Adaptive KL (\textbf{AKL})~\citep{akl}, Token-level Distillation (\textbf{ToDi})~\citep{todi}, and Entropy-Aware On-Policy Distillation (\textbf{EOPD})~\citep{eopd}. We reproduce all baselines using their official implementations and align key training settings.

\begin{figure*}[t]
  \includegraphics[width=\linewidth]{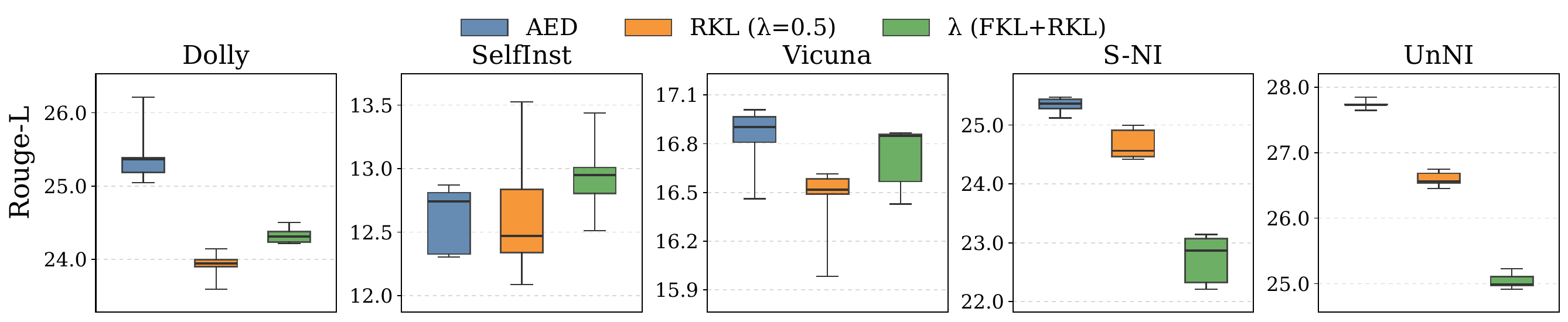}
  \caption{Box plots of ROUGE-L scores for AED and its ablation variants across five evaluation benchmarks.}
  \label{fig:box}
\end{figure*}
\subsection{Main Results}
\label{sec:main_results}

\paragraph{Overall Performance.}
As shown in the Table~\ref{tab:main_results}, the proposed AED consistently achieves favorable overall performance across different model architectures and parameter scales, securing the highest average scores. 

AED surpasses the teacher on several benchmarks, consistent with findings from \citet{bann} and \citet{minillm}. This outperformance likely stems from two factors: (i) the SFT-trained teacher inherently suffers from exposure bias during generation \citep{minillm}; (ii) our adaptive weighting helps the student achieve sharper probability concentration on high-confidence tokens, resulting in cleaner target distributions.

\begin{figure}[h]
  \includegraphics[width=\columnwidth]{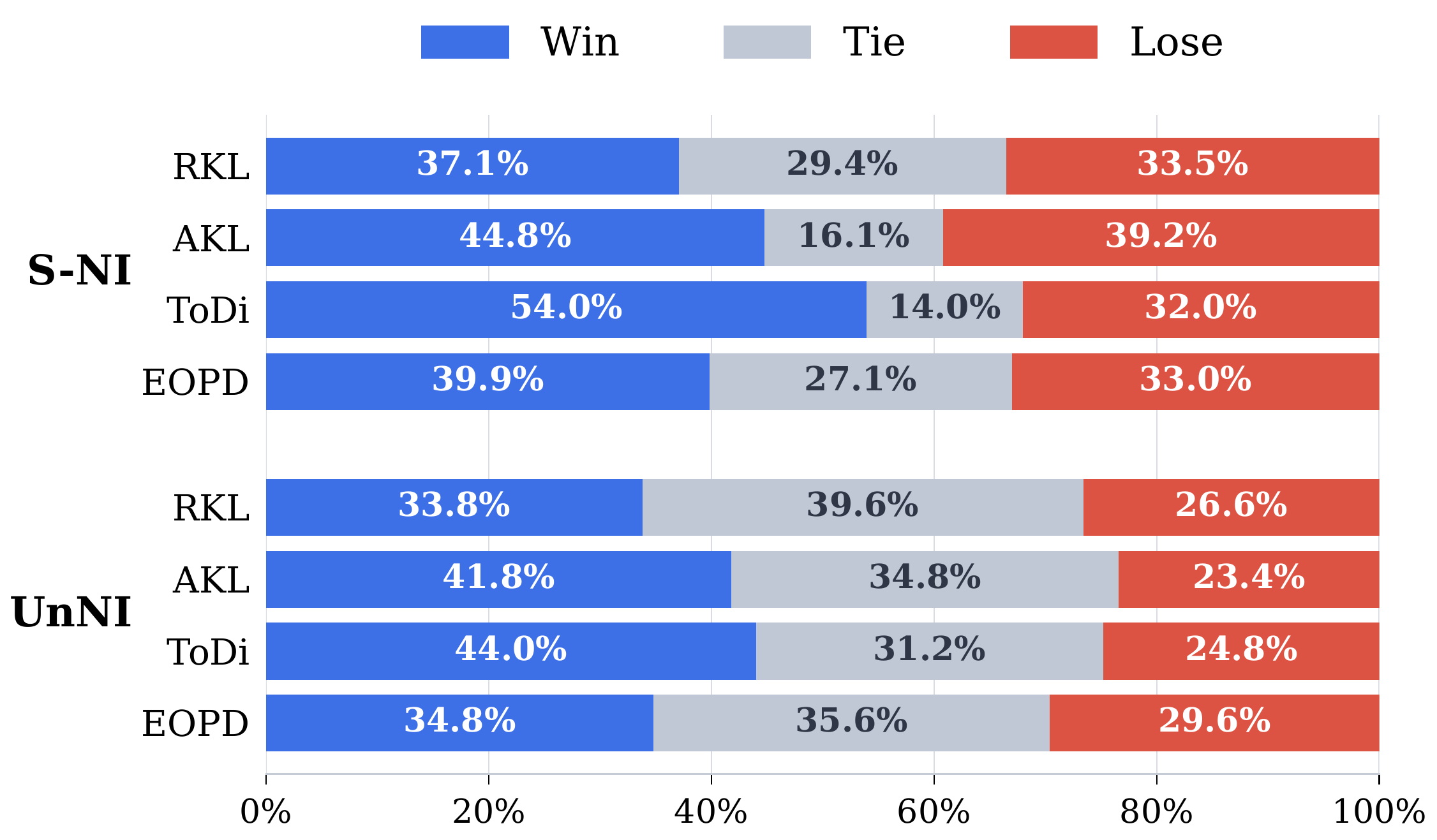}
  \caption{Pairwise preference win-tie-lose rates comparing AED against RKL, AKL, ToDi and EOPD.}
  \label{fig:gpteval}
\end{figure}

\paragraph{LLM-as-a-Judge Evaluation.}
Following the evaluation framework of \citet{alpaca_eval}, we further conduct pairwise preference evaluation using GPT-5.4, accessed via API, as the LLM judge. The evaluation is performed with the TinyLLaMA-1.1B student on the full S-NI test set and the first 500 examples of UnNI. As shown in Figure~\ref{fig:gpteval}, we compare AED against RKL, AKL, ToDi, and EOPD, and report the win, tie, and loss rates of AED. AED achieves higher win rates than loss rates in all comparisons, indicating better response quality under the judge evaluation.

\subsection{Ablation Analysis}
To investigate the specific contribution of our adaptive weighting mechanism, we compare the standard AED, implemented with GPT-2 as the base language model, with two experimental variants: (1) \textbf{RKL ($\lambda=0.5$)}, which removes adaptivity by assigning an equal, static weight of 0.5 to both constituent terms within the objective; (2) \textbf{$\lambda$ (FKL+RKL)}, which utilizes the adaptive $\lambda$ to balance FKL and RKL components while retaining both objectives simultaneously.

As illustrated in Figure~\ref{fig:box}, the standard AED significantly outperforms all variants across the majority of benchmarks. The superiority of AED over the fixed $\lambda=0.5$ baseline demonstrates that a static balance fails to capture the intricate token-level trade-offs between probability concentration and exploration. 

For the $\lambda$ (FKL+RKL) variant, directly applying adaptive weighting to FKL and RKL lacks explicit theoretical motivation and is therefore more heuristic in nature. In addition, the mean-seeking nature of FKL encourages the student to cover the full support of the teacher's distribution. 
In capacity-limited settings, the support-covering pressure of FKL can trade off against concentration on high-confidence teacher modes.

\subsection{Distributional and Entropy Alignment Analysis}
\label{sec:dis_ali}

\begin{table}[h!]
\centering
\small
\setlength{\tabcolsep}{9pt}
\renewcommand{\arraystretch}{1.15}
\begin{tabular}{lcc}
\toprule
\textbf{Methods} & \textbf{RKL} $\downarrow$ & \textbf{MAEG} $\downarrow$ \\
\midrule

\multicolumn{3}{c}{GPT-2 1.5B $\rightarrow$ GPT-2 120M} \\
\midrule
RKL  & 2.9556 & \underline{0.4002} \\
AKL  & 2.7281 & 0.4042 \\
ToDi & 2.6728 & 0.4104 \\
EOPD & \underline{1.3814} & 0.4180 \\
\rowcolor{lightblue}
AED  & \textbf{1.3212} & \textbf{0.3251} \\

\midrule
\multicolumn{3}{c}{LLaMA2 7B $\rightarrow$ TinyLLaMA 1.1B} \\
\midrule
RKL  & 2.6069 & 0.4792 \\
AKL  & 2.5620 & 0.4840 \\
ToDi & 2.9130 & \underline{0.4613} \\
EOPD & \underline{2.5350} & 0.6715 \\
\rowcolor{lightblue}
AED  & \textbf{2.2428} & \textbf{0.4243} \\

\midrule
\multicolumn{3}{c}{Qwen3-4B $\rightarrow$ Qwen3-0.6B} \\
\midrule
RKL  & 2.7257 & \textbf{0.2549} \\
AKL  & 2.5911 & \underline{0.2606} \\
ToDi & 2.5718 & 0.2652 \\
EOPD & \textbf{2.3496} & 0.3161 \\
\rowcolor{lightblue}
AED  & \underline{2.3554} & 0.2747 \\

\bottomrule
\end{tabular}
\caption{RKL and entropy alignment under three teacher--student configurations. \textbf{Bold} and \underline{underlined} values indicate the best and second-best results within each configuration, respectively.}
\label{tab:rkl_ent}
\end{table}

\begin{table*}[h!]
\centering
\small
\setlength{\tabcolsep}{7pt}
\renewcommand{\arraystretch}{1.15}

\begin{tabular}{lcc|cc|cc}
\toprule
\multirow{2}{*}{Method}
& \multicolumn{2}{c|}{MATH-500}
& \multicolumn{2}{c|}{AMC23}
& \multicolumn{2}{c}{AIME 2024} \\
& Avg@8 & Pass@8
& Avg@8 & Pass@8
& Avg@8 & Pass@8 \\
\midrule

RKL
& 14.63\% & 53.40\%
& 9.06\% & \underline{37.50\%}
& 0.42\% & \underline{3.33\%} \\

AKL
& 7.23\% & 34.00\%
& 3.44\% & 20.00\%
& \underline{0.83\%} & \textbf{6.67\%} \\

ToDi
& 9.75\% & 43.40\%
& 8.13\% & 35.00\%
& 0.42\% & \underline{3.33\%} \\

EOPD
& \underline{40.65\%} & \underline{66.00\%}
& \underline{17.50\%} & 35.00\%
& \underline{0.83\%} & \textbf{6.67\%} \\

\rowcolor{lightblue}
AED
& \textbf{47.80\%} & \textbf{71.40\%}
& \textbf{23.75\%} & \textbf{50.00\%}
& \textbf{2.08\%} & \textbf{6.67\%} \\

\bottomrule
\end{tabular}

\caption{Accuracy (\%) results on three mathematical reasoning benchmarks. \textbf{Bold} and \underline{underlined} values indicate the best and second-best results, respectively.}
\label{tab:math_res}
\end{table*}

\begin{figure*}[h!]
  \includegraphics[width=\linewidth]{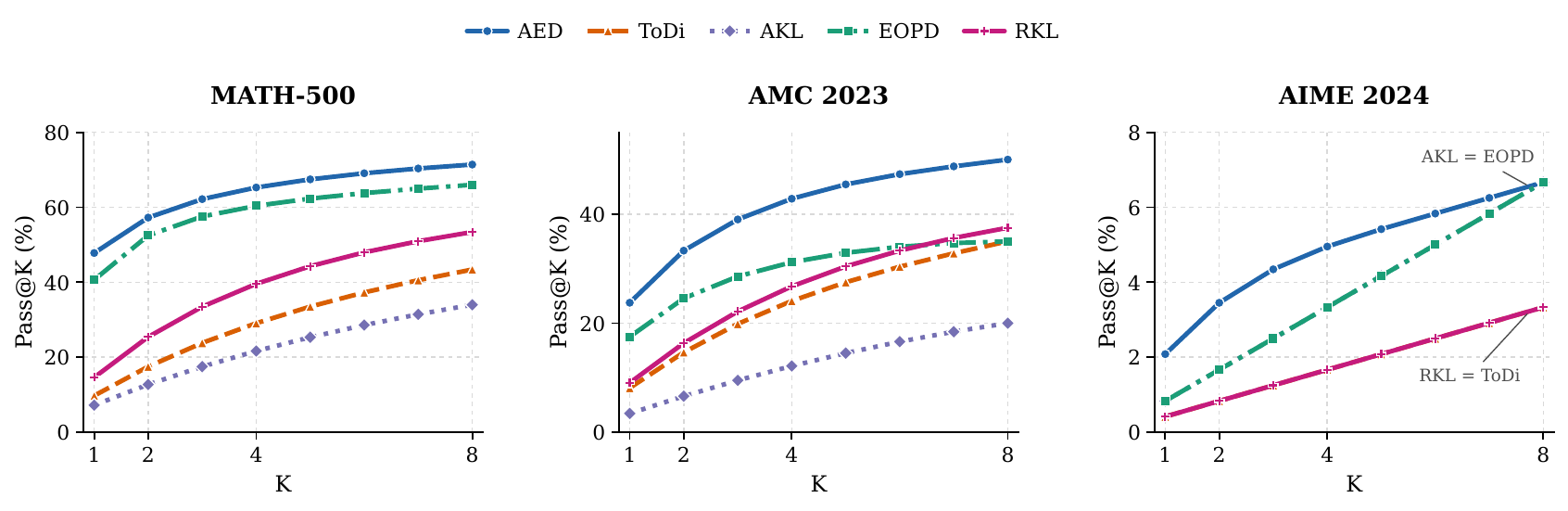}
  \caption{Pass@k results on the evaluated benchmarks.}
  \label{fig:Passk}
\end{figure*}

To conduct a fine-grained analysis, we consider three teacher--student model pairs. For each pair, the teacher generates eight distinct responses for every prompt in the S-NI dataset, and we evaluate the student's predictive distribution against that of the teacher at each token position. We assess distributional alignment using token-level RKL and entropy alignment using the Mean Absolute Entropy Gap (MAEG), defined as the average absolute difference between the teacher and student predictive entropies. Both metrics are averaged over all tokens across the eight responses, with lower values indicating closer teacher--student alignment.

As shown in Table~\ref{tab:rkl_ent}, AED achieves the lowest RKL and MAEG values in the GPT-2 and LLaMA2--TinyLLaMA configurations. Under Qwen3, AED obtains competitive RKL and MAEG alignment with the best-performing methods.
Overall, these results show that entropy-guided adjustment improves the alignment of predictive distributions and uncertainty levels between the student and teacher. Additional alignment results on mathematical reasoning datasets are provided in Appendix~\ref{app:math_dis}.

\subsection{Mathematical Reasoning}
To further evaluate the generalization ability of AED beyond instruction-following tasks, we extend the comparison to mathematical reasoning. We use Qwen3-4B-GRPO \citep{qwen3grpo} as the teacher model and Qwen3-0.6B \citep{qwen3} as the student model. The student models are trained on the MATH \citep{math} training set. We evaluate the resulting student models on MATH-500 \citep{math}, AMC 2023 \citep{amc23} and AIME 2024 \citep{aime2024} using Avg@8 and Pass@8.

As shown in Table~\ref{tab:math_res}, AED achieves the best overall performance across all three benchmarks. 
Figure~\ref{fig:Passk} compares Pass@$k$ for $k \in \{1,2,4,8\}$ across three mathematical reasoning benchmarks. AED achieves higher Pass@1 and consistently higher Pass@$k$ under the same sampling budget, indicating better single-sample accuracy and sampling efficiency.

%% file: chap6_conclu.tex
\section{Conclusion}
In this work, we decompose the RKL objective and derive its equivalence to a tempered RKL framework. Based on this analysis, we propose AED, which adaptively adjusts token-level distillation intensity according to teacher uncertainty. Experiments demonstrate that AED achieves strong performance in task evaluation, distributional alignment, and entropy alignment.

%% file: chap7_limit.tex
\section*{Limitations}
Despite its advantages, certain limitations of our work warrant discussion. First, AED is designed for open-source LLMs and requires vocabulary consistency between the teacher and student to perform token-level alignment. Extending this to black-box APIs or heterogeneous vocabularies remains an open direction. Second, due to computational resource constraints, we did not scale experiments to ultra-large models. Crucially, our theoretical formulations are entirely scale-agnostic, supporting seamless application to LLMs of any capacity when resources are available.

%% file: chap8_app.tex
\appendix

\section{Analytic Optimum of the Token-Level Adaptive Objective}
\label{app:analytic_optimum1}

We derive the analytic optimum of our token-level adaptive loss and demonstrate its mathematical equivalence to minimizing the Reverse RKL divergence between the student distribution and a temperature-scaled teacher distribution. 

\subsection{Lagrangian Optimization for the Adaptive Objective}

Let $p_t(v) = p(y_t=v \mid x, y_{<t})$ and $q_t(v) = q_\theta(y_t=v \mid x, y_{<t})$ denote the teacher and student probabilities over the vocabulary $\mathcal{V}$ at step $t$, respectively. Given a fixed prefix $y_{<t}$, the token-level expected loss to be minimized is defined as:
\begin{equation}
\begin{aligned}
    \mathcal{L}_t(q_t) = \sum_{v \in \mathcal{V}} q_t(v) \Big[ -&\lambda_t \log p_t(v) \\
    &+ (1-\lambda_t) \log q_t(v) \Big].
\end{aligned}
\end{equation}

To find the optimal student distribution $q_t^*$, we minimize $\mathcal{L}_t(q_t)$ subject to the probability simplex constraint $\sum_v q_t(v) = 1$. We construct the Lagrangian:
\begin{equation}
\begin{aligned}
    \mathcal{L}(q_t, \eta)
    &= -\lambda_t \sum_{v \in \mathcal{V}} q_t(v)\log p_t(v) \\
    &\quad + (1-\lambda_t)\sum_{v \in \mathcal{V}} q_t(v)\log q_t(v) \\
    &\quad + \eta \left(\sum_{v \in \mathcal{V}} q_t(v) - 1\right).
\end{aligned}
\end{equation}

Setting the partial derivative with respect to $q_t(v)$ to zero yields:
\begin{equation}
\begin{aligned}
    \frac{\partial \mathcal{L}}{\partial q_t(v)} &= -\lambda_t \log p_t(v) \\
    &\quad + (1-\lambda_t)\big(1 + \log q_t(v)\big) + \eta = 0.
\end{aligned}
\end{equation}

Solving for $\log q_t(v)$, we have:
\begin{equation}
    \log q_t(v) = \frac{\lambda_t}{1-\lambda_t} \log p_t(v) + C,
\end{equation}
where $C = -1 - \frac{\eta}{1-\lambda_t}$ is a constant independent of $v$. Thus, the analytic optimum explicitly takes the form of a scaled distribution:
\begin{equation}
    q_t^*(v) \propto p_t(v)^{\alpha_t}, \quad \text{where} \quad \alpha_t = \frac{\lambda_t}{1-\lambda_t}.
\end{equation}

\subsection{Mathematical Equivalence to Tempered RKL}

This optimality condition reveals that the adaptive objective implicitly minimizes a divergence against a temperature-scaled transformation of the teacher's predictive distribution.
To explicitly establish the connection to the RKL framework, we define the normalized, temperature-scaled (tempered) teacher distribution as:
\begin{equation}
    \tilde{p}_t^{(\alpha_t)}(v) = \frac{p_t(v)^{\alpha_t}}{Z_t(\alpha_t)},  \, \text{where} \, Z_t(\alpha_t) = \sum_{u \in \mathcal{V}} p_t(u)^{\alpha_t}.
\end{equation}
We can now rewrite the original loss $\mathcal{L}_t(q_t)$ using $\alpha_t$:
\begin{equation}
\begin{aligned}
    \mathcal{L}_t(q_t)
    &= (1-\lambda_t)
    \sum_{v \in \mathcal{V}} q_t(v)
    \Big[
    -\alpha_t \log p_t(v) \\
    &\quad + \log q_t(v)
    \Big].
\end{aligned}
\end{equation}

Taking the logarithm of the tempered teacher distribution gives $\alpha_t \log p_t(v) = \log \tilde{p}_t^{(\alpha_t)}(v) + \log Z_t(\alpha_t)$. Substituting this back into the loss yields:
\begin{equation}
\begin{aligned}
    \mathcal{L}_t(q_t)
    &= (1-\lambda_t)\sum_{v \in \mathcal{V}} q_t(v)
    \Bigg[
    \log \frac{q_t(v)}{\tilde{p}_t^{(\alpha_t)}(v)} \\
    &\qquad\qquad
    - \log Z_t(\alpha_t)
    \Bigg] \\
    &= (1-\lambda_t)\sum_{v \in \mathcal{V}} q_t(v)
    \log \frac{q_t(v)}{\tilde{p}_t^{(\alpha_t)}(v)} \\
    &\quad - (1-\lambda_t)\log Z_t(\alpha_t)
    \sum_{v \in \mathcal{V}} q_t(v).
\end{aligned}
\end{equation}

Since $\sum_v q_t(v) = 1$, the first term inside the summation can be naturally expressed as the standard Kullback-Leibler divergence:
\begin{equation}
    \sum_{v \in \mathcal{V}} q_t(v)
    \log \frac{q_t(v)}{\tilde{p}_t^{(\alpha_t)}(v)} =
    D_{\mathrm{KL}}\!\left(q_t \,\|\, \tilde{p}_t^{(\alpha_t)}\right).
\end{equation}

Ultimately, the expected loss objective simplifies to:
\begin{equation}
    \mathcal{L}_t(q_t) = (1 - \lambda_t) \left[ D_{\text{KL}}(q_t \parallel \tilde{p}_t^{(\alpha_t)}) - \log Z_t(\alpha_t) \right].
\end{equation}
This derivation reveals that minimizing our token-level adaptive loss is mathematically equivalent to minimizing the RKL divergence between the student distribution and the tempered teacher distribution. When directly targeting $p_t$, standard RKL can exhibit stronger mode-seeking behavior. Our objective executes a controllable mode-seeking paradigm. By dynamically adjusting the inverse temperature $\alpha_t$ in response to teacher entropy, the objective safely smooths the target distribution in high-uncertainty regions, while preserving confident mode-seeking in low-uncertainty regions.

\section{Implementation Details}
\label{app:imple_details}

\subsection{Training details}
\paragraph{Instruction Following.} We use the \texttt{databricks/dolly-15k} dataset for training, which contains 11K training samples, 1K validation samples, and 500 test samples. All training processes are conducted on a server equipped with two NVIDIA RTX A800 (80GB) GPUs. The training cost is approximately 3 GPU hours for GPT-2, 30 GPU hours for TinyLLaMA, and 21.5 GPU hours for Qwen3.
Detailed training configurations for the three model architectures are summarized in Table~\ref{tab:ins_hyperparameters}.

\begin{table}[h!]
\centering
\scriptsize
\setlength{\tabcolsep}{1.5pt}
\begin{tabular}{lccc}
\toprule
\textbf{Parameter}
& \textbf{GPT-2 (120M)}
& \textbf{TinyLLaMA (1.1B)}
& \textbf{Qwen3 (0.6B)} \\
\midrule
LoRA              & No                  & Yes                 & Yes                 \\
Optimizer         & AdamW               & AdamW               & AdamW               \\
LR                & $2 \times 10^{-5}$ & $5 \times 10^{-6}$ & $5 \times 10^{-6}$ \\
LR scheduler      & Cosine              & Cosine              & Cosine              \\
Mini-batch size   & 32                  & 16                  & 16                  \\
Batch size        & 32                  & 32                  & 32                  \\
Max response length  & 256     & 256         & 256 \\
Temperature       & 1.0                 & 1.0                 & 1.0                 \\
Top-$p$           & 1.0                 & 1.0                 & 1.0                 \\
Epochs            & 3                   & 10                  & 10                  \\
\bottomrule
\end{tabular}
\caption{Instruction-following hyperparameters.}
\label{tab:ins_hyperparameters}
\end{table}

\paragraph{Mathematical Reasoning.}
We use the \texttt{verl-team/lighteval-MATH-preprocessed} dataset for training, which contains 7.5K training samples and 5K test samples. All training processes are conducted on a server equipped with two NVIDIA A800 (80GB) GPUs. The training cost is approximately 3.93 GPU hours for Qwen3. Detailed training configuration is summarized in Table~\ref{tab:math_hyperparameters}.

\begin{table}[h!]
\centering
\small
\setlength{\tabcolsep}{12pt}
\renewcommand{\arraystretch}{1.10}

\begin{tabular}{lc}
\toprule
\textbf{Parameter} & \textbf{Qwen3 (0.6B)} \\
\midrule
LR        & $5 \times 10^{-6}$ \\
LR scheduler    & Cosine \\
Optimizer            & AdamW \\
Batch size  & 128 \\
Mini-batch size      & 32 \\
Top-$p$              & 1.0 \\
Max response length  & 1024 \\
Temperature & 1.0 \\
Epochs      & 3 \\
\bottomrule
\end{tabular}
\caption{Mathematical-reasoning hyperparameters.}
\label{tab:math_hyperparameters}
\end{table}

\subsection{Evaluation details}
\paragraph{Instruction Following.}
For generation-based evaluation, we sample model responses with temperature set to $1.0$, \texttt{top-k} set to $0$, and \texttt{top-p} set to $1.0$. 
For the evaluation datasets, we use the full test sets of DollyEval, SelfInst, Vicuna, and S-NI, while for UnNI we evaluate on the first 10,000 examples.
To account for randomness in generation, we conduct evaluations with five random seeds, $\{10, 20, 30, 40, 50\}$, and report the mean ROUGE-L score together with the standard deviation across these runs. The final training checkpoint is used for evaluation.

\paragraph{Mathematical Reasoning.}
For mathematical reasoning evaluation, we generate eight responses for each problem with a maximum response length of 2,048 tokens. Sampling is performed with a temperature of $1.0$, \texttt{top-k} of $0$, and \texttt{top-p} of $0.8$. We report Avg@8, the average accuracy over the eight responses, and Pass@8, the proportion of problems for which at least one response is correct. The final training checkpoint is used for evaluation.

\section{Distributional and Entropy Alignment in Mathematical Reasoning}
\label{app:math_dis}

\begin{table}[h!]
\centering
\small
\setlength{\tabcolsep}{10pt}
\renewcommand{\arraystretch}{1.15}
\begin{tabular}{lcc}
\toprule
\textbf{Methods} & \textbf{RKL} $\downarrow$ & \textbf{MAEG} $\downarrow$ \\
\midrule
\multicolumn{3}{c}{Qwen3-4B-GRPO $\rightarrow$ Qwen3-0.6B} \\
\midrule
RKL  & 0.3283 & 0.1502 \\
AKL  & 0.3592 & 0.1627 \\
ToDi & 0.3590 & 0.1622 \\
EOPD & \underline{0.2504} & \underline{0.0987} \\
\rowcolor{lightblue}
AED  & \textbf{0.2473} & \textbf{0.0870} \\
\bottomrule
\end{tabular}
\caption{Distributional and entropy alignment for mathematical reasoning distillation on AMC23.}
\label{tab:math_rkl_ent}
\end{table}

We further evaluate token-level RKL and entropy alignment on the AMC23 dataset using the MAEG. Except for the dataset, the evaluation settings follow those in Section \ref{sec:dis_ali}. The results show that AED obtains the best alignment.